\def\year{2019}\relax
\documentclass[letterpaper]{article} 
\usepackage{aaai19}  
\usepackage{times}  
\usepackage{helvet}  
\usepackage{courier}  
\usepackage[hyphens]{url}  

\usepackage{amsmath, amsfonts, amssymb}
\usepackage{graphicx}
\usepackage{algorithm}
\usepackage{algorithmic}
\usepackage{bbm}
\usepackage{ulem}

\usepackage[colorinlistoftodos, textwidth=17mm]{todonotes}
\usepackage{color}
\usepackage{enumitem}
\usepackage[breaklinks]{hyperref}
\hypersetup{breaklinks=true}

\setlist[itemize]{itemsep=0.5pt, wide=\parindent}
\setlist[enumerate]{itemsep=0.5pt, wide=\parindent}

\DeclareMathSizes{10}{8}{6}{6}

\def\pp{\mathbb{P}}
\def\rr{\mathbb{R}}

\def\nn{\mathbb{N}}

\def\mcD{\mathcal{D}}
\def\mcI{\mathcal{I}}

\def\one{\mathbbm{1}}

\nocopyright

\begin{document}

\title{Multi-Task Determinantal Point Processes for Recommendation}

\author{Romain Warlop \\
fifty-five \& \\ SequeL Team, Inria Lille-Nord Europe \\
romain@fifty-five.com
\And 
J\'{e}r\'{e}mie Mary \\ 
Criteo AI Lab \\
j.mary@criteo.com
\And
Mike Gartrell \\ 
Criteo AI Lab \\
m.gartrell@criteo.com
}

\maketitle


\begin{abstract}
Determinantal point processes (DPPs) have received significant attention in the
recent years as an elegant model for a variety of machine learning tasks, due to
their ability to elegantly model set diversity and item quality or popularity.
Recent work has shown that DPPs can be effective models for product
recommendation and basket completion tasks. We present an enhanced DPP model
that is specialized for the task of basket completion, the multi-task DPP. We
view the basket completion problem as a multi-class classification problem, and
leverage ideas from tensor factorization and multi-class classification to
design the multi-task DPP model.  We evaluate our model on several real-world datasets,
and find that the multi-task DPP provides significantly better predictive quality than a
number of state-of-the-art models.
\end{abstract}

\section{Introduction}
Increasing the number of items in the average shopping basket is a major concern for online
retailers. While there are a wide range of possibles strategies, this work
focuses on the algorithm responsible for proposing a set of items that is best suited to completing
the current shopping basket of the user.

Basket analysis and completion is a very old task for machine
learning. For many years association rule
mining~\cite{Agrawal:1993:MAR:170036.170072} 
has been the state-of-the-art. Even though there are different variants of this
algorithm, the main principle involves computing the conditional probability of
buying an additional product by counting co-occurrences in past observations.
Due to computational cost and robustness, modern approaches favor item-to-item
collaborative filtering~\cite{1167344}, or using logistic regression to predict if a
user will purchase an item based on binary purchase scores obtained from shopping
baskets~\cite{Lee:2005:CCF:1707421.1707525}.

As reported in the \textit{Related Work} section, standard collaborative
filtering approaches need to be extended to correctly capture diversity among products.
Practitioners often mitigate this problem by adding constraints to the
recommended set of items. As an example, when using categorical information, it
is possible to force the recommendation of a pair of matching shoes when trousers are
added to the basket, even if natural co-sale patterns would lead to the recommendation
of other trousers. In this situation the presence of diversity in the
recommendations is not directly driven by the learning algorithm, but by side
information and expert knowledge.  Ref.~\cite{Teo:2016:APD:2959100.2959171}
proposes an effective Bayesian method for learning the weights of the categories
in the case of visual search when categories are known.

Sometimes we need to learn diversity without relying on extra information. Naive
learning of diversity directly from the data without using side information
comes at a high computational cost, because the number of possible sets
(baskets) grows exponentially with the number of items. The issue is not
trivial, even when we want to be able to add only one item to an existing set,
and becomes even harder when we want to add more than one item with the idea of
maximizing the diversity of the final recommended set.

Refs.~\cite{DBLP:conf/aaai/GartrellPK17,DBLP:conf/recsys/GartrellPK16} address this
combinatorial problem using a model based on Determinantal Point Processes
(DPPs) for basket completion. DPPs are elegant probabilistic models of repulsion
from quantum physics, which are used for a variety of tasks in machine
learning~\cite{Kulesza:2012:DPP:2481023}.
They allow sampling a diverse set of points, with similarity and popularity encoded
using a positive semi-definite matrix called the kernel. Efficient algorithms
for marginalization and conditioning DPPs are available. From a practical
perspective, learning the DPP kernel is a challenge because the associated
likelihood is non-convex, and learning it from observed sets of items is
conjectured to be NP-hard~\cite{Kulesza:2012:DPP:2481023}.

For basket completion it is natural to consider that sets are the baskets which
converted to sales. In this setting, the DPP is parameterized by a kernel matrix
of size $p \times p$, where $p$ is the size of the catalog. Thus the number of
parameters to fit grows quadratically with $p$, and the computational complexity
for learning, prediction, and sampling grows cubicly with $p$. As learning a
full-rank DPP is hard,~\cite{DBLP:conf/aaai/GartrellPK17} proposes regularizing
the DPP by constraining the kernel to be low rank. This regularization also
improves generalization and offers more diversity in recommendations, without
hurting predictive performance. In fact in many settings the predictive quality
is also improved, making the DPP a very desirable tool for modeling baskets.
Moreover, the low-rank assumption also enables substantially better runtime
performance compared to a full-rank DPP.

Nevertheless, because of the definition of the DPP, as described in the
\textit{Model} section, this low-rank assumption for the kernel means that any
possible baskets with more items than the chosen rank will receive a probability
estimate of 0. This approach is thus impossible to use for large baskets, and
some other regularizations of the DPP kernel may be more appropriate. The
contributions of this paper are fourfold:
\begin{itemize}
	\item We modify the constraints over the kernel to support large baskets.
    \item We model the probability over all baskets by adding a logistic
    function on the determinant computed from the DPP kernel. We adapt the
    training procedure to handle this nonlinearity, and evaluate our model
    on several real-world basket datasets.
    \item By leveraging tensor factorization, we propose a new way to regularize
    the kernel among a set of tasks. This approach also leads to enhanced
    predictive quality.
    \item We show that this new model, which we call the multi-task DPP, allows
    us to capture directed basket completion.  That is, we can leverage the
    information regarding the order in which items are added to a cart to
    improve predictive quality.
 \end{itemize}
Furthermore, we show that these ideas can be combined for further improvements
to predictive quality, allowing our multi-task DPP model to outperform
state-of-the-art models by a large margin.

We begin by introducing our proposed algorithm, and then proceed to evaluate its
effectiveness on several real-world datasets. We then discuss
related work before concluding and introducing possible future work. 
\section{Model}
\label{sec:model}
\subsection{Background}
Determinantal Point Processes (DPPs) were originally used  to model a
distribution over particles that exhibit a repulsive
effect~\cite{Vershik2001book}. Recently, interest in leveraging this repulsive
behavior has led to DPPs receiving increasing attention within the machine
learning community.  Mathematically, discrete DPPs are distributions over
discrete sets of points, or in our case items, where the model assigns a
probability to observing a given set of items. Let $\mcI$ denote a set of items,
and $L$ the kernel matrix associated with the DPP whose entries encode item
popularity and the similarity between items. The probability of observing the
set $\mcI$ is proportional to the determinant of the principal submatrix of $L$
indexed by the items in $\mcI$: $\pp(\mcI) \propto \det L_{\mcI}$~\footnote{To
define a probability measure on the DPP, the normalization factor is $\det(L +
I)$, because $\sum_{\mcI} \det L_{\mcI} = \det (L+I)$.}. Thus, if $p$ denotes
the number of items in the item catalog, the DPP is a probability measure on
$2^p$ (the power set, or set of all subsets of $p$). The kernel $L$ encodes item
popularities and the similarities between items, where the diagonal entry
$L_{ii}$ represents the popularity of item $i$, and the off-diagonal entry
$L_{ij} = L_{ji}$ represents the similarity between items $i$ and $j$. A determinant can
be seen as a volume from a geometric viewpoint, and therefore more diverse sets
will tend to have larger determinants.
For example, 
the probability of selecting two items $i$ and $j$ together can be computed as
\begin{eqnarray}
\label{dpp_intuition}
\pp[\{i,j\}] \propto \begin{vmatrix}
L_{ii} & L_{ij} \\ L_{ji} & L_{jj} 
\end{vmatrix} = L_{ii} L_{jj}-L_{ij}^2
\end{eqnarray}
In equation~\ref{dpp_intuition} we can see that the more similar $i$ and $j$
are, the less likely they are to be sampled together. The definition of the
entries $L_{ij}$ will therefore determine the repulsive behavior of the kernel
for the task. For instance, if similarity is defined using image descriptors,
then images of differing appearance will be selected by a DPP. On the other
hand, if the entries $L_{ij}$ are learned using previously observed sets, such as
e-commerce baskets~\cite{DBLP:conf/aaai/GartrellPK17}, then co-purchased items $i$ and
$j$ are likely to be sampled by the DPP, and thus the "similarity" $L_{ij}$
will be low. In an application such as a search engine or in document summarization, the
kernel may be defined using feature descriptors $\psi_i \in \rr^D$ (i.e tf-idf
of the text), and a relevance score $q_i \in \rr^+$ of each item $i$ such that
$L_{ij} = q_i \psi_i^T \psi_j q_j$, which favors relevant items (large $q_i$) and
discourages lists composed of similar items. 

\subsection{Logistic DPP}
Our objective is to find a set of items that are most likely to be purchased
together. We formulate this as a classification problem, where the goal is to
predict if a specific set of items will generate a conversion from the user,
which we denote as $Y \in \{0,1\}$. We model the class label $Y$ as a Bernoulli random
variable with parameter $\phi(\mcI)$, where $\mcI$ is the set of items and $\phi$
is a function that we will define below:
\begin{equation}
p(y|\mcI) = \phi(\mcI)^y(1-\phi(\mcI))^{1-y}
\end{equation} 
We model the function $\phi$ using a DPP. 

We assume that there exists a latent space such that diverse items in this space
are likely to be purchased together. Similarly
to~\cite{DBLP:conf/aaai/GartrellPK17}, we introduce a low-rank factorization of
the kernel matrix $L \in \rr^{p \times p}$:
\begin{equation}
L = VV^T+D^2
\end{equation}
where $V \in \rr^{p \times r}$ is a latent matrix where each row vector $i$
encodes the $r$ latent factors of item $i$.  $D$ is a diagonal matrix that, and
together with $||V_i||$, represents the intrinsic quality or popularity of each
item. The squared exponent on $D$ insures that we always have a valid positive
semi-definite kernel. We then define $\phi(\mcI) \propto \det (V_{\mcI,:}
V_{\mcI,:}^T+D^2) \geq 0$. Note that without the diagonal term, the choice of
$r$ would restrict the cardinality of the observable set, because $|\mcI| > r$
would imply $\phi(\mcI) = 0$ when $D\equiv0$. Using this term will ensure that
the success probability of any set will be positive, but the cross-effects will
be lower for sets of cardinality higher than $r$. We also see that items with
similar latent vectors are less likely to be sampled than items with different
latent vectors, since similar vectors will produce a parallelotope with a
smaller volume. To normalize the probability and encourage separation between
vectors we use a logistic function on $\Phi$ such that:
\begin{eqnarray}
\label{themodel}
\phi(I) = \pp(y=1|\mcI) & \doteq & 1-\exp(-w \det L_{\mcI}) \\
& \doteq & \sigma(w \det L_{\mcI})
\end{eqnarray}
Usually the logistic function is of the form $1/(1+\exp(-w \det L_{\mcI}))$.
However, in our case the determinant is always positive, since $L$ is positive
semi-definite, which would result in $\pp(y=1|\mcI)$ always greater than 0.5
with such a function. By construction, our formulation allows us to obtain a
probability between 0 and 1. Finally, $w \in \rr$ is a scaling parameter, to
be chosen by cross-validation, that insures that the exponential does not
explode, since the diagonal parameter will be approximately 1. \\
\textbf{Learning}. In order to learn the matrix $V$ we assume the existence of
historical data $\{\mcI_m,y_m\}_{1 \leq m \leq M}$, where $\mcI_m$ is a set of
items, and $y_m$ is a label set to 1 if the set has been purchased, and 0
otherwise. This training data allows us to learn the matrices $V$ and $D$ by maximizing the
log-likelihood of the data. To do so, we first write the click probability for
all $y$ as
\begin{equation}
\label{themodel_y}
\pp(y|\mcI) = \sigma(w \det L_{\mcI})^y (1-\sigma(w \det L_{\mcI}))^{1-y} 
\end{equation}
The log-likelihood $f(V,D)$ can then be written as
\begin{eqnarray*}
\label{loglikelihood_ST}
f(V,D) & = & \log \prod_{m=1}^M \pp(y_m|\mcI_m) - \frac{\alpha_0}{2} \sum_{i=1}^p \alpha_i (||V_i||^2 + ||D_i||^2) \\
& = & \sum_{m=1}^M \log \pp(y_m|\mcI_m) - \frac{\alpha_0}{2} \sum_{i=1}^p \alpha_i (||V_i||^2 + ||D_i||^2)
\end{eqnarray*}
Following \cite{DBLP:conf/aaai/GartrellPK17}, $\alpha_i$ is an item
regularization weight that is inversely proportional to item popularity.
The matrices $V$ and $D$ are learned by maximizing the log-likelihood
using stochastic gradient ascent. Details on the optimization algorithms and the
gradient equations are available in the supplementary material.

\subsection{Multi-task DPP}
We now propose a modification to the previously introduced model that is better
suited for the basket completion task. To do so we enhance the logistic DPP for
the basket completion scenario, where we model the probability that the user
will purchase a specified additional item based on the items already present in the
user's shopping basket. We formulate this as a multi-task classification
problem, where the goal is to predict whether the user will purchase a given
target item based on the user's basket. In this setting there are as many tasks
as there are items in the catalog, $p$ (minus the items already in the basket).
Learning one kernel per task would be impossible in practice and suffer from
sparsity issues. Indeed, with one kernel per task, each target item would be
present in only a fraction of the baskets, and thus dramatically reduce the size
of the training set per kernel. To solve this issue we utilize a low-rank
tensor. We use a cubic tensor $K \in \rr^{p \times p \times p}$, where each
slice $\tau$ (noted $K_{\tau}$) of $K$ is the task (low-rank) kernel. By
assuming that the tensor $K$ is low-rank, we are able to implement sharing of
learned parameters between tasks, as shown in the following equation:
\begin{equation}
\label{rescal_decomposition}
K_{\tau} = V R_{\tau}^2 V^T + D^2
\end{equation}
where $V \in \rr^{p \times r}$ are the item latent factors that are common to
all tasks, and $R_{\tau} \in \rr^{r \times r}$ is a task specific matrix that
models the interactions between the latent components of each task. In order to
balance the degrees of freedom between tasks and items, we further assume that
$R_{\tau}$ is a diagonal matrix. Therefore, the diagonal vector of $R_{\tau}$
models the latent factors of each task, and the latent factors of the item can
be seen as the relevance of the product for each latent factor. As is the case
for the matrix $D$, the squared exponent on $R_{\tau}$ ensures that we always
have a valid kernel. This decomposition is similar to the RESCAL
decomposition~\cite{ICML2011Nickel_438}, without the additional bias term and a
diagonal constraint on the slice specific matrix. We also use a different
learning procedure due to the use of the logistic function. The probability that
a set of items $\mcI$ will be successful for task
$\tau$ is
\begin{equation}
\label{themodel_multitask}
\pp(y_{\tau}=1|\mcI) = \sigma(w \det K_{\tau,\mcI})= 1-\exp(-w \det K_{\tau,\mcI}) 
\end{equation}
%
Therefore, the log-likelihood $g(V,D,R) \doteq g$ is
\begin{eqnarray*}
\label{loglikelihood_MT}
g = \sum_{m=1}^{M} \log \pp(y_{\tau}|\mcI_m) - \frac{\alpha_0}{2} \sum_{i=1}^p \alpha_i (||V_i||^2 + ||D_i||^2 + ||R^i||^2)
\end{eqnarray*}
where each observation $m$ is associated with a task, and $\mcI_m$ is the set of
items associated with an observation.  As previously described, matrices $V$,
$D$, and $(R_{\tau})_{\tau \in \{1,\cdots,p\}}$ are learned by maximizing the
log-likelihood using stochastic gradient ascent. Details on the optimization
algorithms and the gradient equations are available in the supplementary
material.

\section{Experiments}
We evaluate the performance of our model on the basket completion problem on several real-world datasets,
and compare to several state-of-the-art baselines. \\
\\
\noindent
\textbf{MODELS}  
\begin{itemize}
\item \textbf{Our models}. To understand the impact of the different components
of our model compared to the low-rank DPP model, we evaluated the following
versions of our model:
\begin{itemize}
\item \textbf{\textsc{Logistic DPP}}: This version of our model is
similar to the low-rank DPP model, with the addition of the logistic function.
To determine what item to recommend we use a greedy approach, where we select
the next item such that the probability of the basket completed with this item
is the largest. We used $w=0.01$.
\item \textbf{\textsc{Multi-task log-DPP without bias}}:
In this version of the model we set $D\equiv 0$, which allows us to measure the
impact of capturing the item bias in a separate matrix.  The matrix $V$ encodes
the latent factors of items present in the basket, while each matrix $R_{\tau}$
encodes the latent factors of each target item $\tau$ that can be added to a basket. We used $w=0.01$.
\item \textbf{\textsc{Multi-task log-DPP}}: This is the full version of our
model, with bias enabled. We used $w=0.01$.
\end{itemize}
Our datasets do not provide explicit negative information.  To generate negative
feedback for our models we created negatives targets from observed baskets by
sampling a random item among those items not in the basket. This approach could
be improved through better negative sampling strategies, but since this is not
part of our primary contributions we leave this investigation for future work. 
\item \textbf{Baselines}. The primary goal of our work is to improve
state-of-the-art results provided by DPPs and introduce new modeling
enhancements to DPPs. However, for the sake of completeness we also compare with
other strong baseline models provided by state-of-the-art collaborative
filtering approaches.
\begin{itemize}
\item \textbf{\textsc{Poisson Factorization}}
(PF)~\cite{DBLP:journals/corr/GopalanHB13} is a probabilistic matrix
factorization model generally used for recommendation applications with implicit
feedback. Since our datasets contain no user id information, we consider each
basket to be a different user, and thus there are as many users as baskets in
the training set. In practice this can cause issues with high memory
consumption, since the number of baskets can be very large. 
\item \textbf{\textsc{Factorization Machines}}
(FMs)~\cite{Rendle:2010:FM:1933307.1934620} is a general approach that models
$d$th-order interactions using low-rank assumptions. FMs are usually used with
$d=2$, since this corresponds to classic matrix factorization and because
complexity increases linearly with $d$. Additionally, there is no open-source FM
implementation that supports $d>2$. For these reasons, we use $d=2$ in our
experiments. As with PF, to learn the FM model we consider each basket as a
unique user. For fairness in comparison with our models, we also tried FM with
negative sampling based on item popularity. However, we did not see any
substantial improvement in model performance when using this negative sampling
approach. 
\item \textbf{\textsc{Low-Rank DPP}}~\cite{DBLP:conf/aaai/GartrellPK17} is a
low-rank DPP model, suitable for basket completion, where the
determinant of the submatrix of the kernel corresponds to the probability that
all the items will be bought together in a basket.
\item \textbf{\textsc{Bayesian Low-Rank
DPP}}~\cite{DBLP:conf/recsys/GartrellPK16} is the Bayesian version of the
low-rank DPP model.
\item \textbf{\textsc{Associative Classifier}} (AC) is an algorithm that
computes the support of a purchased set of items in order to obtain completion
rules. As in~\cite{DBLP:conf/aaai/GartrellPK17}, we used the Classification
Based on Associations (CBA) algorithm~\cite{Liu:1998:ICA:3000292.3000305},
available at~\cite{arules}, with minimum support of 1.0\% and maximum
confidence thresholds of 20.0\%. Unlike other models, AC does not provide
estimates for all possible sets. Therefore, we cannot compute results for some
metrics used in our evaluation, such as MPR (described below). 
\item \textbf{\textsc{Recurrent Neural Network}} This RNN model~\cite{DBLP:journals/corr/HidasiKBT15} 
is adapted for session-based recommender systems. The RNN requires ordered
sequences, and thus we only evaluate this model on the Instacart dataset
(described below), where the order in which items were added to each basket is
available. We use the implementation of this model available from~\cite{RNNcode}.
\end{itemize}
\end{itemize}
For all models we tried different hyperparameter settings, such as the
number of latents factors and regularization strength, and report the best results here.
In the interest of reproducibility, all code used for our experiments is available at
ANONYMOUS\_REF.
\\
\\
\noindent
\textbf{DATASETS.} For our basket completion experiments we use the following four datasets. 
The first three ones contains undirected baskets, that is there is no notion of the order in 
which the items have been added to the basket, whereas the last contains directed basket, that 
is ordered sets:
\begin{itemize}
\item \textbf{Amazon Baby Registries} is a public dataset consisting of
$110,006$ registries and $15$ disjoint registry categories. For the purposes of
comparison with~\cite{DBLP:conf/recsys/GartrellPK16}, we perform two
experiments.  The first experiment is conducted using the diaper category, which
contains 100 products and approximately 10,000 baskets, composed of 2.4
items per basket on average.  The second experiment is performed on the
concatenation of the diaper, apparel, and feeding categories (sometimes noted
here as D.A.F for Diaper+Apparel+Feedings), which contains 300 products and
approximately 17,000 baskets, composed of 2.6 items per basket on average.
The item categories are disjoint; for example, no basket containing diaper
products will contain apparel products. This concatenation of disjoint
categories can present difficulties for classic matrix factorization
models~\cite{DBLP:conf/recsys/GartrellPK16}, which may prevent these models from
learning a good embedding of items.
\item \textbf{Belgian Retail Supermarket} is a public
dataset~\cite{brijs99using} that contains 88,163 sets of items that have been
purchased together, with a catalog of 16,470 unique items. Each basket
contains 9.6 items on average. AC cannot be trained on this dataset because
this approach does not scale to large item catalogs.
\item \textbf{UK retail dataset} is a public dataset~\cite{Chen2012} that
contains 22,034 sets of items that have been purchased together, among a
catalog of 4,070 unique items.  This dataset contains transactions from a
non-store online retail company that primarily sells unique all-occasion gifts,
and many customers are wholesalers. Each basket contains 18.5 items on
average, with a number of very large baskets.  Modeling these large baskets
requires using a very large number of latent factors for the low-rank DPP,
leading to somewhat poor results for this model. This is not an issue for our
model, due to the item bias that is captured in a separate matrix. However, for
purposes of comparison, we removed all baskets containing more than 100 items
from this dataset; note that the low-rank DPP still requires 100 latent
factors to model these baskets. AC could not be trained on this dataset because
it does not scale to large item catalogs.
\item \textbf{Instacart} is, to the best of our knowledge, the only public dataset 
\footnote{https://www.instacart.com/datasets/grocery-shopping-2017} 
that contains the order in which products were added to baskets. It is composed
of three datasets containing online grocery shopping behavior for more than
200,000 Instacart users: a ``train" dataset, a ``test" dataset, and a ``prior"
dataset. We use only the ``train" dataset in our experiments, and remove items
that appear less than 15 times and baskets of size lower than 3. This
results in a dataset containing 700,052 sets of items and 10,531 unique
items.
\end{itemize}

\noindent
\textbf{METRICS}. To evaluate the performance of each model we compute the Mean
Percentile Rank and precision@$K$ for $K=5, 10,$ and $20$:
\begin{itemize}
\item \textbf{Mean Percentile Rank (MPR)}: Given a basket $B$, we compute the
percentile rank $\text{PR}_{i_B}$ of the held-out item, $i_B$. Let
$p_i \doteq P(Y=1|B)$.  Then 
\begin{equation}
\text{PR}_{i_B} = \frac{\sum_{i=1}^p \one(p_{i_B} \geq p_i)}{p} \times 100\%
\end{equation}
The MPR is the average PR over all baskets in the test set:
\begin{equation}
\text{MPR} = \frac{\sum_{B \in \mathcal{T}} \text{PR}_{i_B}}{|\mathcal{T}|}
\end{equation}
where $\mathcal{T}$ is the set of all baskets in the test set. A MPR of 100\%
means that the held-out item always receives the highest predictive score, while
a MPR of 50\% corresponds to a random sorting. Higher MPR scores are better.
\item \textbf{Precision@$K$} is the fraction of test baskets where the held-out
item is in the top $K$ ranked items. 
\begin{equation}
\text{precision@}K = \frac{\sum_{B \in \mathcal{T}} \one(\text{rank}_{i_B} \leq K)}{|\mathcal{T}|}
\end{equation}
Higher precision@$K$ scores are better. 
\end{itemize}

We evaluated the predictive quality of our models for both undirected and directed
basket completion. Recall that for undirected baskets, there is no information
regarding the order in which items are added to baskets, while directed baskets
do contain such ordering structure. We use the Amazon, Belgian retail, and UK 
retail datasets for our undirected basket experiments, while the Instacart dataset 
is used for our directed basket experiments. For all experiments we use a random split
of 70\% of the data for training, and 30\% for testing.

\subsection{Results for Undirected Baskets}
For our undirected basket experiments,  we remove one item at random from each
basket in the test set. We then evaluate the model prediction according to the
predicted score of this removed item using the metrics described below. 

\noindent
Looking at Table~\ref{tab:results}, we see that classic collaborative filtering
models sometimes have difficulty providing good recommendations in the
basket-completion setting. Perhaps more surprising, but already described
in~\cite{DBLP:conf/aaai/GartrellPK17}, is that for the Amazon datasets, PF
provides MPR performance that is approximately equivalent to a random model. For
the Amazon diaper dataset this poor performance may be a result of the small
size of each basket (around 2.4 items per basket on average), thus each "user"
is in a cold start situation, and it is therefore difficult to provide good predictions.
Poor performance on the diaper+apparel+feedings Amazon's dataset may result from
the fact that, apart from the small basket size of 2.61 items on average, this
dataset is composed of three disjoint categories. These disjoint categories can
break the low-rank assumption for matrix factorization-based models, as
discussed in~\cite{DBLP:conf/recsys/GartrellPK16}. This issue is somewhat
mitigated in FM, due to the integration of an item bias into the model. This
item bias allows the model to capture item popularity and thus provide
acceptable performance in some cases.

Finally, the DPP-based models generally
outperform the FM model.  This is likely due to the fact that DPP models
are able to capture higher-order interactions within baskets, while FM
is only able to capture second-order interactions, since $d=2$ for this model.
\\
\noindent
\textbf{Low Rank DPP vs Multi-task DPP}.
We now turn to a performance comparison between our primary baseline, the
low-rank DPP model, and our multi-task DPP model. From Table~\ref{tab:results},
we see that our approaches provide a substantial increase in performance for
both Amazon datasets, with relative improvements between 10\% and 70\%. One
factor that accounts for this performance improvement is that unlike the
low-rank DPP, which models the probability that a set of items will be bought
together, our approach directly models the basket completion task. In our
multi-task DPP model, the extra dimensions allow the model to capture the
correlation between each item in the basket and the target item, as well as the
global coherence of the set.

Regarding the three-category Amazon dataset, a good model should not be impacted
by the fact the all of the three categories are disjoint. Therefore, the
precision@$K$ scores should be approximately the same for both the
single-category and three-category datasets, since we observe similar
performance for each category independently. Since the MPR is 78\% for one
category, the MPR on the three-category dataset should be approximately 93\%,
since on average for each category the target item is in the 22nd position
over the 100 items in the single-category catalog. Therefore, the target item should be in the
22nd position over the 300 items in the three-category catalog, resulting in a MPR of $1-22/300 = 93\%$.
Our models come close to these numbers, but still exhibit some small degradation
for the three-category dataset. Finally, we note that for this dataset, a model
that samples an item at random from the right category would have a
precision@20 of 20\%, since there are 100 items per category. The low-rank
DPP model provides close to this level of performance. Taken together, these
observations indicate our model is robust to the disjoint category problem, and
explains the 70\% relative improvement we see for our model on the
precision@20 metric. On the UK Retail dataset the improvements of our
algorithm are still substantial for precision@$K$, with a relative improvement of
between 20\% and 30\% (MPR is down by 5\%). We also observe the same
decrease in MPR for our logistic DPP model, but precision@$K$ is similar to the
low-rank DPP. On the Belgian Retail dataset we see that all models
provide similar performance. For this dataset, baskets come from an offline
supermarket, where it is possible that customers commonly purchased similar
products at specific frequencies. Consequently it may be easy to capture
frequent associations between purchased items, but very difficult to discover
more unusual associations, which may explain why all models provide
approximately the same performance.
\\
\noindent
\textbf{Logistic DPP vs Multi-task DPP}.
To better understand the incremental performance of our model, we focus on the
results of the logistic DPP and the multi-task log-DPP models. We see that
the logistic (single-task) model does not improve over the low-rank DPP on average,
indicating that the logistic component of the model does not contribute to
improved performance. However, we argue that this formulation may still be
valuable in other classification applications, such as those with explicit
negative feedback. For the multi-task log-DPP model, we see that the version of
this model without bias is responsible for almost all of the performance
improvement.  Some additional lift is obtained when capturing the item
popularity bias in a separate matrix. Since most of the gain comes from the
multi-task kernel, one may ask if we could use the multi-task kernel without the
logistic function and obtain similar results. We believe that this is not the
case for two reasons. First, since we are clearly in a classification setting,
it is more appropriate to use a logistic model that is directly tailored for
such applications. Second, without the logistic function, each slice of the
tensor should define a probability distribution, meaning that the probability of
purchasing an additional product should sum to one over all possible baskets.
However, we could add an arbitrarily bad product that would never be purchased,
resulting in a probability of zero for buying that item in any basket,
which would break the distributional assumption. 

\renewcommand{\arraystretch}{0.95}
\begin{table*}
\small
\begin{center}
\begin{tabular}{ccccccc}
\hline
model & dataset & $r$ & MPR & Prec.@$5$ & Prec.@$10$ & Prec.@$20$ \\ \hline
\textsc{Associative Classifier} & Amazon (diaper) & - & - & 16.66 & 16.66 & 16.66 \\
\textsc{Poisson Factorization} & Amazon (diaper) & 40 & 50.30 & 4.78 & 10.03 & 19.90 \\
\textsc{Factorization Machines} & Amazon (diaper) & 5 & 67.92 & 24.01 & 32.62 & 46.25 \\
\textsc{Low Rank DPP} & Amazon (diaper) & 30 & 71.65 & 25.48 & 35.80 & 49.98 \\
\textsc{Bayesian Low Rank DPP} & Amazon (diaper) & 30 & 72.38 & 26.31 & 36.21 & 51.51 \\
\textsc{Logistic DPP} & Amazon (diaper) & 50 & 71.08 & 23.7 & 34.01 & 48.44 \\ 
\textsc{multi-task logDPP no bias} & Amazon (diaper) & 50 & 77.5 & 32.7 & 45.77 & 61.00 \\ 
\textbf{\textsc{multi-task logDPP}} & Amazon (diaper) & 50 & \textbf{78.41} & \textbf{34.73} & \textbf{47.42} & \textbf{62.58} \\ \hline

\textsc{Associative Classifier} & Amazon (D.A.F) & - & - & 4.16 & 4.16 & 4.16 \\
\textsc{Poisson Factorization} & Amazon (D.A.F) & 40 & 51.36 & 4.16 & 5.88 & 9.08 \\
\textsc{Factorization Machines} & Amazon (D.A.F) & 5 & 65.21 & 10.62 & 16.71 & 24.20 \\
\textsc{Low Rank DPP} & Amazon (D.A.F) & 30 & 70.10 & 13.10 & 18.59 & 26.92 \\
\textsc{Bayesian Low Rank DPP} & Amazon (D.A.F) & 30 & 70.55 & 13.59 & 19.51 & 27.83 \\
\textsc{Logistic DPP} & Amazon (D.A.F) & 60 & 69.61 & 12.65 & 19.8 & 27.86 \\ 
\textsc{multi-task logDPP no bias} & Amazon (D.A.F) & 60 & 88.77 & 18.33 & 28.00 & 43.57 \\ 
\textbf{\textsc{multi-task logDPP}} & Amazon (D.A.F) & 60 & \textbf{89.80} & \textbf{20.53} & \textbf{30.86} & \textbf{45.79} \\ \hline

\textsc{Poisson Factorization} & Belgian Retail Supermarket & 40 & 87.02 & 21.46 & 23.06 & 23.90 \\
\textsc{Factorization Machines} & Belgian Retail Supermarket & 10 & 65.08 & 20.85 & 21.10 & 21.37 \\
\textsc{Low Rank DPP} & Belgian Retail Supermarket & 76 & 88.52 & 21.48 & 23.29 & 25.19 \\
\textsc{Bayesian Low Rank DPP} & Belgian Retail Supermarket & 76 & 89.08 & 21.43 & 23.10 & 25.12 \\
\textsc{Logistic DPP} & Belgian Retail Supermarket & 75 & 87.35 & 21.17 & 23.11 & 25.77 \\ 
\textsc{multi-task logDPP no bias} & Belgian Retail Supermarket & 75 & 87.42 & 21.02 & 23.35 & 25.13 \\ 
\textsc{multi-task logDPP} & Belgian Retail Supermarket & 75 & 87.72 & 21.46 & 23.37 & 25.57 \\ \hline

\textsc{Poisson Factorization} & UK Retail & 100 & 73.12 & 1.77 & 2.31 & 3.01 \\
\textsc{Factorization Machines} & UK Retail & 5 & 56.91 & 0.47 & 0.83 & 1.50 \\
\textsc{Low Rank DPP} & UK Retail & 100 & \textbf{82.74} & 3.07 & 4.75 & 7.60 \\
\textsc{Bayesian Low Rank DPP} & UK Retail & 100 & 61.31 & 1.07 & 1.91 & 3.25 \\
\textsc{Logistic DPP} & UK Retail & 100 & 75.23 & 3.18 & 4.99 & 7.83 \\ 
\textsc{multi-task log DPP no bias} & UK Retail & 100 & 77.67 & 3.82 & 5.98 & 9.11 \\ 
\textbf{\textsc{multi-task logDPP}} & UK Retail & 100 & 78.25 & \textbf{4.00} & \textbf{6.20} & \textbf{9.40} \\ 
\hline
\end{tabular}
\vspace{-0.1in}
\caption{Result of all models on all datasets. $r$ denotes the number of latent factors. Best results within each dataset are in bold.}
\label{tab:results}
\end{center}
\vspace{-0.1in}
\end{table*}


\begin{table}
\small
\begin{center}
\begin{tabular}{cccccc}
\hline
model & $\mathcal{P}$ & MPR & P@$5$ & P@$10$ & P@$20$ \\ 
\hline
\textsc{FM} & (1) & 61.10 & 4.55 & 6.3 & 7.67 \\
\textsc{Low Rank DPP} & (1) & 76.46 & 7.37 & 8.07 & 9.23 \\
\textsc{Multi-Task DPP} & (1) & 80.46 & 4.62 & 7.23 & 10.51 \\
\hline
\textsc{FM} & (2) & 62.47 & 9.35 & 10.66 & 11.92 \\
\textsc{Low Rank DPP} & (2) & 61.16 & 7.49 & 8.05 & 8.8 \\
\textsc{RNN} & (2) & 73.31 & 1.08 & 1.99 & 3.2 \\
\textsc{Multi-Task DPP} & (2) & \textbf{90.07} & \textbf{9.91} & \textbf{13.67} & \textbf{19.97} \\
\hline
\textsc{Multi-Task DPP} & (3) & 80.65 & 5.23 & 6.05 & 9.72
\end{tabular}
\vspace{-0.1in}
\caption{Performance of the models on Instacart dataset for the three protocols $\mathcal{P}$. 
All models used $80$ latent factors except FM that used $5$ latent factors.}
\label{tab:instacartresults}
\end{center}
\vspace{-0.25in}
\end{table}

\subsection{Results for Directed Baskets}
Recall that directed baskets contain ordering information which 
may provide substantial information for basket completion. 
In order to evaluate the ability of our model to capture directed basket completions,
we performed three experimental protocols on the Instacart dataset, which
contains ordered sequences of items added to baskets. Each protocol varies in the way that we
remove the item to predict from the basket:
\begin{enumerate}
	\item As with previous experiments, we remove one item at random from each
	basket. For the low-rank DPP and FM models this item removal is performed
	only for baskets in the test set. We do not remove items from baskets in the
	training set, since these baskets are used to learn inter-item correlation
	patterns that are applied to new baskets. For the multi-task DPP, we perform
	item removal for both the training and test sets. Item removal in the
	training set is appropriate for the multi-task DPP since the removed item
	corresponds to the target item for this model.
	\item We remove the last item added to each basket. For the low-rank DPP and
	FM models, this is done only for the test set, and for the multi-task DPP
	this done for both the training and test sets. Since we consider ordered
	sequences, we also evaluate the RNN model using this protocol, where item
	removal is done only for the test set.
	\item We remove one item at random from each basket in the training set, and
	the last item added to each basket in the test set. Here we evaluate only
	the performance of the multi-task DPP model, since it is the only model that
	involves item removal in the training set.
\end{enumerate}

\noindent
Looking at Table~\ref{tab:instacartresults} and comparing the multi-task DPP
results for protocols (2) and (3), we see that our model performs much better
when predicting the last item added to a basket when training is also done by
removing the last item added (protocol (3)). This allows us to conclude that the
order in which items are added to the basket is important, otherwise both
protocols would give similar results. Next, when comparing the results of
protocols (1) and (2), we see that multi-task DPP performance is lower when the
model is trained to predict a randomly removed item than when trained to predict
the last item added, while we see that this pattern is reversed for the low-rank
DPP. This indicates that the low-rank DPP, although well suited to modeling item
co-occurrence probabilities, is unable to capture directed basket completion.
Finally, we see, surprisingly, that the RNN model does not provide good
performance for this task. This relatively poor performance may come from the
fact that the basket lengths are too small in this dataset for the RNN to learn
the item sequences within baskets correctly. 



\section{Related Work}
The topics of DPPs, basket completion, and diversity have significant attention
in recent years.

In addition to the previously discussed work, DPPs have been used for natural
language processing in order to discover diverse threads of
documents~\cite{Gillenwater:2012:DDS:2390948.2391026}, and to enhance diversity
in recommender
systems~\cite{DBLP:journals/corr/abs-1709-05135}. 
Unlike in our
application where we learn the kernels, in these applications the kernel is
constructed using previously obtained latent factors, for instance using tf-idf
~\cite{Gillenwater:2012:DDS:2390948.2391026}. These latent factors are scaled by
a relevance score learned in a more conventional fashion. For example, these
relevance scores may represent the predicted rating of a particular user, or the
similarity between the text in a document and the user query. Ultimately, these
applications involve sampling from the DPP specified by this kernel, where the
kernel parameters trade off between relevance and diversity. However, sampling
from such a DPP efficiently is difficult, and this has lead to work on different
sampling techniques. Ref.~\cite{NIPS2014_5564} 
relies on MCMC sampling, while~\cite{DBLP:journals/corr/abs-1709-05135}
proposes a greedy solution based on Cholesky decomposition. 

Several algorithms have been proposed for learning the DPP kernel matrix.
Ref.~\cite{NIPS2014_5564} uses an expectation-maximization (EM) algorithm to
learn a non-parametric form of the DPP kernel matrix.
Ref.~\cite{DBLP:journals/corr/MarietS15} proposes a fixed-point algorithm called
Picard iteration, which is much faster than EM, but still slower
than~\cite{DBLP:conf/aaai/GartrellPK17}. Bayesian learning methods have also
been proposed to learn the DPP
kernel~\cite{DBLP:conf/recsys/GartrellPK16,DBLP:conf/icml/AffandiFAT14}.

Improving diversity in recommender systems has also been studied without the use
of DPPs, including, among other work,
~\cite{Christoffel:2015:BWA:2792838.2800180,PuthiyaParambath:2016:CAR:2959100.2959149,Vargas:2014:ISD:2645710.2645744}. 
For instance,~\cite{Christoffel:2015:BWA:2792838.2800180} relies on random walk
techniques to enhance diversity.
In~\cite{PuthiyaParambath:2016:CAR:2959100.2959149}, the authors propose trading
off between the relevance of the recommendation and diversity by introducing a
coverage function to force the algorithm to produce recommendations that cover
different centers of the interests of each user.
Finally, the authors
of~\cite{Vargas:2014:ISD:2645710.2645744} propose transforming the problem of
recommending items to users into recommending users to items. They introduce a
modification of nearest-neighbor methods, and a probabilistic model that allows
isolation of the popularity bias and favors less popular items. 

Regarding basket completion, associative classifiers have long been the
state-of-the-art~\cite{Agrawal:1993:MAR:170036.170072}, 
despite requiring
very heaving computational load for training, and manual tuning for key
parameter choices such as lift and confidence thresholds. Later work focuses on
the task of purchase prediction by adapting collaborative filtering methods.
Ref.~\cite{Mild2003} proposes a solution based on nearest-neighbor models,
while~\cite{Lee:2005:CCF:1707421.1707525} relies on binary logistic regression
to predict if a user will purchase a given item. More recently,
DPPs~\cite{DBLP:conf/aaai/GartrellPK17,DBLP:conf/recsys/GartrellPK16} may now be
considered among the class of models belonging to the new state-of-the-art for
basket completion, in light of their effectiveness both in terms of accuracy and
training speed. Finally, classic collaborative filtering models tailored for
positive and unlabelled
data~\cite{Hu:2008:CFI:1510528.1511352,DBLP:journals/corr/GopalanHB13} 
may be effectively used for basket completion.

\section{Conclusion and Future Work}
In this paper we have proposed an extension of the DPP model that leverages
ideas from multi-class classification and tensor factorization. While our model
can be applied to a number of machine learning problems, we focus on the problem
of basket completion. We have shown through experiments on several datasets
that our model provides significant improvements in predictive quality compared
to a number of competing state-of-the-art approaches and can appropriately capture 
directed basket completion. In future work we plan to
investigate other applications of our model, such as user conversion prediction,
attribution, and adversarial settings in games.  We also plan to investigate
better negative sampling methods for positive-only and unlabelled data. Finally,
we also plan to investigate other types of loss functions, such as hinge loss,
and other types of link functions for DPPs, such as the Poisson function, to
tailor DPPs for regression problems. We believe that this work will allow 
us to customize DPPs so that they are suitable for many additional applications.

\clearpage
\Urlmuskip=0mu plus 1mu\relax
\bibliographystyle{aaai}
\bibliography{biblio} 

\clearpage
\appendix

\section{Appendix}
\subsection{Logistic DPP} 
Recall that the logistic DPP log-likelihood is:
{\small
\begin{eqnarray*}
\label{loglikelihood_ST}
f(V,D) & = & \log \prod_{m=1}^M \pp(y_m|\mcI_m) - \frac{\alpha_0}{2} \sum_{i=1}^p \alpha_i (||V_i||^2 + ||D_i||^2) \\
& = & \sum_{m=1}^M \log \pp(y_m|\mcI_m) - \frac{\alpha_0}{2} \sum_{i=1}^p \alpha_i (||V_i||^2 + ||D_i||^2)
\end{eqnarray*}}
\noindent
\textbf{Optimization}. We maximize the log-likelihood using stochastic gradient
ascent with Nesterov's Accelerated Gradient, which is a form of momentum. 
To simplify notation, we define $[m] \doteq \mcI_m$ and $\sigma_m = \sigma(w \det
L_{[m]})$. Let $i \in \{1, \cdots, p \}, k \in \{1, \cdots, r\}$. 
\\
\noindent
\textbf{lemma} When $D$ is fixed, the gradient of (\ref{loglikelihood_ST}) with respect to $V_{ik}$ is
\begin{eqnarray}
\label{ST_Vgradient}
\frac{\partial f}{\partial V_{ik}} & = & 2 w \sum_{m, i \in [m]} ([L_{[m]}^{-1}]_{:,i} V_{:,k}) \frac{y_m-\sigma_m}{\sigma_m} \det L_{\mcI_m} \nonumber \\ 
& & - \alpha_0 \alpha_i V_{ik} 
\end{eqnarray}

\textbf{Proof} Without the regularization term we have
{\small
\begin{eqnarray}
\frac{\partial f}{\partial V_{ik}} & = & \sum_{m, i \in [m]} \frac{y_m}{\sigma_m} \frac{\partial \sigma_m}{\partial V_{ik}}+ \frac{1-y_m}{1-\sigma_m}\left(-\frac{\partial \sigma_m}{\partial V_{ik}} \right) \\
& = & w \sum_{m, i \in [m]} \frac{y_m-\sigma_m}{\sigma_m} \frac{\partial \det L_{[m]}}{\partial V_{ik}} \\
& = & w \sum_{m, i \in [m]} \frac{y_m-\sigma_m}{\sigma_m} \ \text{tr}\left(L_{[m]}^{-1} \frac{\partial L_{[m]}}{\partial V_{ik}}\right) \det L_{\mcI_m} \\
& = & 2 w \sum_{m, i \in [m]} ([L_{[m]}^{-1}]_{:,i} V_{:,k}) \frac{y_m-\sigma_m}{\sigma_m} \det L_{\mcI_m} \label{gradient_last_step}
\end{eqnarray}}
where (\ref{gradient_last_step}) follows from
\begin{equation}
\left[\frac{\partial L_{[m]}}{\partial V_{ik}}\right]_{s,t} = V_{sk} \delta_{i,t} + V_{tk} \delta_{i,s}
\end{equation}
Therefore,
\begin{eqnarray}
\text{tr}\left(L_{[m]}^{-1} \frac{\partial L_{[m]}}{\partial V_{ik}}\right) & = & \sum_{s,t} ( V_{sk} \delta_{i,t} + V_{tk} \delta_{i,s}) [L_{[m]}^{-1}]_{s,t} \\
& = & \sum_s [L_{[m]}^{-1}]_{s,i} V_{sk} + \sum_t [L_{[m]}^{-1}]_{i,t} V_{tk} \nonumber \\
& = & 2 \sum_s [L_{[m]}^{-1}]_{s,i} V_{sk} 
\end{eqnarray}
adding the derivative of the regularization term concludes the proof. $\square$

\noindent
\textbf{lemma} When $V$ is fixed, the gradient of (\ref{loglikelihood_ST}) with respect to $D_i$ is
\begin{eqnarray}
\label{ST_Dgradient}
\frac{\partial f}{\partial D_{ii}} & = & 2 w \sum_{m, i \in [m]} ([L_{[m]}^{-1}]_{i,i} D_{i,i}) \frac{y_m-\sigma_m}{\sigma_m} \det L_{[m]} \nonumber \\
& & - \alpha_0 \alpha_i D_{ii} 
\end{eqnarray}

\noindent
\textbf{Proof} As shown previously, and without the regularization term, we have
{\small
\begin{eqnarray}
\frac{\partial f}{\partial V_{ik}} = w \sum_{m, i \in [m]} \frac{y_m-\sigma_m}{\sigma_m} \ \text{tr}\left(L_{[m]}^{-1} \frac{\partial L_{[m]}}{\partial D_{i,i}}\right) \det L_{\mcI_m} 
\end{eqnarray}}
Since,
\begin{eqnarray*}
\left[\frac{\partial L_{[m]}}{\partial D_{i,i}}\right]_{s,t} & = & 2 D_{i,i} \delta_{s,i} \delta_{t,i} \\
\text{tr}\left(L_{[m]}^{-1} \frac{\partial L_{[m]}}{\partial D_{i,i}}\right) & = & 2 [L^{-1}_{[m]}]_{i,i} D_{i,i}
\end{eqnarray*}
adding the  derivative of the regularization term concludes the proof. $\square$


\subsection{Multi-Task DPP} 

Recall that the multi-task DPP log-likelihood is:
{\small
\begin{eqnarray*}
\label{loglikelihood_MT}
g = \sum_{m}^{M} \log \pp(y_{\tau}|[m]) - \frac{\alpha_0}{2} \sum_{i=1}^p \alpha_i (||V_i||^2 + ||D_i||^2 + ||R^i||^2)
\end{eqnarray*}}
\noindent
\textbf{Optimization}. Since each observation $m$ is attached to a task, we denote $\tau_m$ as the task
that corresponds to observation $m$. Thus we have $\sigma_{m} = \sigma(\det
K_{\tau_m,[m]})$. When there is no ambiguity, we also denote $K_{[m]} \doteq
K_{\tau_m,[m]}$. Let $i \in \{1, \cdots, p \}, k \in \{1, \cdots, r\}$. 
\\
\noindent
\textbf{lemma} When $D$ and $R$ are fixed, the gradient of (\ref{loglikelihood_MT}) with respect to $V_{ik}$ is
\begin{eqnarray}
\label{MT_Vgradient}
\frac{\partial g}{\partial V_{ik}} & = & 2 w \sum_{m, i \in [m]} \frac{y_{\tau_m}-\sigma_m}{\sigma_m} R_{\tau_m,k,k}^2 [K^{-1}_{\tau_m,[m]}]_{:,i} \nonumber \\
& & \cdot V_{:,k}  \det K_{\tau_m,[m]}  - \alpha_0 \alpha_i V_{ik} 
\end{eqnarray}

\noindent
\textbf{Proof} Without the regularization term we have
{\small
\begin{eqnarray}
\frac{\partial g}{\partial V_{ik}} & = & \sum_{m, i \in [m]} \frac{y_{\tau_m}}{\sigma_m} \frac{\partial \sigma_m}{\partial V_{ik}}+ \frac{1-y_{\tau_m}}{1-\sigma_m}\left(-\frac{\partial \sigma_m}{\partial V_{ik}} \right) \\
& = & w \sum_{m, i \in [m]} \frac{y_{\tau_m}-\sigma_m}{\sigma_m} \frac{\partial \det K_{[m]}}{\partial V_{ik}} \\
& = & w \sum_{m, i \in [m]} \frac{y_m-\sigma_m}{\sigma_m} \ \text{tr}\left(K_{[m]}^{-1} \frac{\partial K_{[m]}}{\partial V_{ik}}\right) \\ 
& & \cdot \det K_{[m]}
\end{eqnarray}}
Since,
\begin{eqnarray*}
\label{kernel_entry}
[K_{\tau}]_{s,t} - D^2_{s,t} & = & \sum_{j=1}^r [V R_{\tau}^2]_{s,j} V_{t,j} = \sum_{j=1}^r V_{s,j} R_{\tau,j,j}^2 V_{t,j}
\\
\left[\frac{\partial K_{[m]}}{\partial V_{ik}}\right]_{s,t} & = & R_{\tau,k,k}^2 (V_{t,k} \delta_{s,i}+V_{s,k} \delta_{t,i})
\end{eqnarray*}
Thus, 
\begin{eqnarray*}
\text{tr}\left(K_{[m]}^{-1} \frac{\partial K_{[m]}}{\partial V_{ik}}\right) & = & 2 R_{\tau,k,k}^2 \sum_{s \in [m]} [K^{-1}_{\tau_m,[m]}]_{s,i}  V_{s,k} 
\end{eqnarray*}
adding the regularization term concludes the proof. $\square$

\noindent
\textbf{lemma} When $V$ and $R$ are fixed, the gradient of (\ref{loglikelihood_MT}) with respect to $D_{i,i}$ is
\begin{eqnarray}
\label{MT_Dgradient}
\frac{\partial g}{\partial D_{i,i}} & = & 2 w \sum_{m, i \in [m]} \frac{y_{\tau_m} - \sigma_m}{\sigma_m} [K^{-1}_{t_m,[m]}]_{i,i} D_{i,i} \det K_{\mcI_m} \nonumber \\
& & - \alpha_0 \alpha_i D_{ii} 
\end{eqnarray}

\noindent
\textbf{Proof} Similarly, without the regularization term, we have
{\small
\begin{eqnarray}
\frac{\partial g}{\partial D_{i,i}} =  w \sum_{m, i \in [m]} \frac{y_m-\sigma_m}{\sigma_m} \ \text{tr}\left(K_{[m]}^{-1} \frac{\partial K_{[m]}}{\partial D_{i,i}}\right) \det K_{[m]} 
\end{eqnarray}}
Using (\ref{kernel_entry})
\begin{equation}
\left[\frac{\partial K_{[m]}}{\partial D_{i,i}}\right]_{s,t} = 2 D_{i,i} \delta_{s,i} \delta_{t,i}
\end{equation}
thus
{\small
\begin{eqnarray}
\text{tr}\left(K_{[m]}^{-1} \frac{\partial K_{[m]}}{\partial D_{i,i}}\right) & = & 2 [K^{-1}_{t_m,[m]}]_{i,i} D_{i,i}
\end{eqnarray}}
adding the regularization term concludes the proof. $\square$

\begin{algorithm}[t]
    \begin{algorithmic}
    \STATE \textbf{Input:} $\alpha_0 \in \rr$, $\beta \in \rr$ the momentum coefficient, $m \in \nn$ the minibatch size, $\varepsilon \in \rr$ the gradient step, $t=0$ the iteration counter, $T=0$, past data $\mcD = \{\mcI_m,y_m\}_{1 \leq m \leq M}$.
    \STATE \textbf{Initialization:} Compute item popularity and output regularization weights $\alpha_i$. 
    \STATE Set $D_0 \sim \mathcal{N}(1,0.01)$ on the diagonal and $\tilde{D}_0 \equiv 0$ the gradient accumulation on $D$.
    \STATE Set $V_0 \sim \mathcal{N}(0,0.01)$ everywhere and $\tilde{V}_0 \equiv 0$ the gradient accumulation on $V$.
    \IF{multitask}
    \STATE Set $R_{\tau,0} \sim \mathcal{N}(1,0.01)$ on the diagonal for each task and $\tilde{R}_{\tau,0} \equiv 0$ the gradient accumulation on $R_{\tau}$.
    \ENDIF
    \WHILE{not converged}
    \IF{$m (t+1) > M (T+1)$}
    \STATE Shuffle $\mcD$ and set $T = T +1 $
    \ENDIF 
    \IF{multitask}
    \STATE Update $\left(\tilde{V}_{t+1}, \tilde{D}_{t+1}, (\tilde{R}_{\tau,t+1})_{\tau}\right) = \beta \left(\tilde{V}_t,\tilde{D}_t,(\tilde{R}_{\tau,t})_{\tau} \right) + (1-\beta) \varepsilon \bigtriangledown g(V_t+\beta \tilde{V}_t, D_t+\beta \tilde{D}_t, (R_{\tau,t}+\beta \tilde{R}_{\tau,t})_{\tau})$ according to formulas (\ref{MT_Vgradient}), (\ref{MT_Dgradient}) and (\ref{MT_Rgradient})
    \ELSE
    \STATE Update $(\tilde{V}_{t+1}, \tilde{D}_{t+1}) = \beta (\tilde{V}_t,\tilde{D}_t) + (1-\beta) \varepsilon \bigtriangledown f(V_t+\beta \tilde{V}_t, D_t+\beta \tilde{D}_t)$ according to formulas (\ref{ST_Vgradient}) and (\ref{ST_Dgradient})
    \ENDIF
    \STATE Update $V_{t+1} = V_t + \tilde{V}_{t+1}$
    \STATE Update $D_{t+1} = D_t + \tilde{D}_{t+1}$ 
    \IF{multitask}
    \STATE Update $R_{\tau,t+1} = R_{\tau,t} + \tilde{R}_{\tau,t+1}$ for all $\tau$
    \ENDIF
    \ENDWHILE
    \end{algorithmic}
    \caption{Optimization algorithm.}
    \label{alg:multitask}
    \end{algorithm}

\noindent
\textbf{lemma} When $V$ and $D$ are fixed, the gradient of (\ref{loglikelihood_MT}) with respect to $R_{k,k}$ is
\begin{eqnarray}
\label{MT_Rgradient}
\frac{\partial g}{\partial R_{\tau,k,k}} & = & 2 w \sum_{m, \tau \in [m]} \frac{y_{\tau} - \sigma_m}{\sigma_m} \ R_{\tau,k,k} {K^{-1}}_{[m]} \cdot V_{:,k} \nonumber\\
& & \cdot V_{:,k}^T  \det K_{[m]} - \alpha_0 \alpha_{\tau} R_{\tau,k,k,} 
\end{eqnarray}

\noindent
\textbf{Proof} Similarly, without the regularization term, we have
{\small
\begin{eqnarray}
\frac{\partial g}{\partial R_{\tau,k,k}} = w \sum_{m, i \in [m]} \frac{y_{\tau}-\sigma_m}{\sigma_m} \ \text{tr}\left(K_{[m]}^{-1} \frac{\partial K_{[m]}}{\partial R_{\tau,k,k}}\right) \det K_{[m]} \\
\end{eqnarray}}
Using (\ref{kernel_entry})
\begin{equation}
\left[\frac{\partial K_{[m]}}{\partial R_{\tau,k,k}}\right]_{s,t} = 2 R_{\tau,k,k} V_{s,k} V_{t,k}
\end{equation}
we obtain
{\small
\begin{eqnarray}
\text{tr}\left(K_{[m]}^{-1} \frac{\partial K_{[m]}}{\partial R_{\tau,k,k}}\right) & = & 2 R_{\tau,k,k} \sum_{s,t \in [m]} [K^{-1}_{[m]}]_{s,t} V_{s,k} \\ 
& & \cdot V_{t,k}
\end{eqnarray}}
adding the regularization term concludes the proof. $\square$

\end{document}